\documentclass{article}

\usepackage{arxiv}

\usepackage[utf8]{inputenc} 
\usepackage[T1]{fontenc}    
\usepackage{hyperref}       
\usepackage{url}            
\usepackage{booktabs}       
\usepackage{nicefrac}       
\usepackage{microtype}      
\usepackage{lipsum}

\usepackage{natbib}

\usepackage{url}
\usepackage{amsmath}
\usepackage{amssymb}

\usepackage[dvips]{graphicx}
\usepackage{psfrag}
\usepackage{epsfig}

\usepackage[symbol]{footmisc}

\title{Computational model discovery\\ with reinforcement learning}

\author{
  Maxime~Bassenne\thanks{Authors contributed equally.}\\
  Laboratory of Artificial Intelligence\\ in Medicine and Biomedical Physics\\
  Stanford University\\
  Stanford, CA 94305\\
  \texttt{bassenne@stanford.edu} \\
   \And
  Adri\'an~Lozano-Dur\'an$^*$\\
  Center for Turbulence Research\\
  Stanford University\\
  Stanford, CA 94305\\
  \texttt{adrianld@stanford.edu} \\
}

\begin{document}
\maketitle



\section{Motivation and objectives}

5 to 0. This is the score by which AlphaStar--a computer program
developed by the artificial intelligence (AI) company DeepMind--beat a
top professional player in Starcraft II, one of the most complex video
games to date \citep{alphastarblog}. This accomplishment tops the list
of sophisticated human tasks at which AI is now performing at a human
or superhuman level \citep{AIreport2019}, and enlarges the previous
body of achievements in game playing
\citep{silver2017mastering,silver2018general} and in other tasks,
e.g. medical diagnoses
\citep{esteva2017dermatologist,raumviboonsuk2019deep,nagpal2019development}.

Our focus is on computational modeling, which is of paramount
importance to the investigation of many natural and industrial
processes. Reduced-order models alleviate the computational
intractability frequently associated with solving the exact
mathematical description of the full-scale dynamics of complex
systems.  The motivation of this study is to leverage recent
breakthroughs in AI research to unlock novel solutions to important
scientific problems encountered in computational science. Following
the post-game confession of the human player defeated by AlphaStar:
``The agent demonstrated strategies I hadn't thought of before, which
means there may still be new ways of playing the game that we haven’t
fully explored yet" \citep{alphastarblog}, we inquire current AI
potential to discover scientifically rooted models that are free of
human bias for computational physics applications.

Traditional reduced-order modeling approaches are rooted in physical
principles, mathematics, and phenomenological understanding, all of
which are important contributors to the human interpretability of the
models. However these strategies can be limited by the difficulty of
formalizing an accurate reduced-order model, even when the exact
governing equations are known. A notorious example in fluid mechanics
is turbulence: the flow is accurately described in detail by the
Navier-Stokes equations; however, closed equations for the large-scale
quantities remain unknown, despite their expected simpler dynamics.
The work by \citet{Jimenez2020} discusses in more extent the
role of computers in the development of turbulence theories and
models.

To address the human intelligence limitations in discovering
reduced-order models, we propose to supplement human thinking with
artificial intelligence. Our work shares the goal of a growing body of
literature on data-driven discovery of differential equations,
which aims at substituting or combining traditional modeling
strategies with the application of modern machine learning algorithms
to experimental or high-fidelity simulation data
(see \citep{brunton2019machine} for a review on machine learning and fluid mechanics). Early work used symbolic regression and evolutionary
algorithms to determine the dynamical model that best describes
experimental data
\citep{bongard2007automated,schmidt2009distilling}. An alternative,
less costly approach consists of selecting candidate terms in a
predefined dictionary of simple functions using sparse regression
\citep{brunton2016discovering,rudy2017data,schaeffer2017learning,wu2019learning}. \citet{raissi2017physics} and \citet{raissi2018hidden} use Gaussian processes to learn unknown
scalar parameters in an otherwise known partial differential
equation. A recent method exploits the connection between neural
networks and differential equations \citep{chen2018neural} to
simultaneously learn the hidden form of an equation and its numerical
discretization
\citep{long2018pde,long2019pde}. \citet{atkinson2019data} use genetic
programming to learn free-form differential equations. The closest
work to ours in spirit is that of \cite{lee2019coarse}, in which the
authors identify coarse-scale dynamics from microscopic observations
by combining Gaussian processes, artificial neural networks, and
diffusion maps. It is also worth mentioning the distinct yet
complementary work by \citet{bar2019learning}, which deals with the
use of supervised learning to learn discretization schemes.

Our three-pronged strategy consists of learning (i) models expressed
in analytical form, (ii) which are evaluated \emph{a posteriori}, and
(iii) using exclusively integral quantities from the reference
solution as prior knowledge. In point (i), we pursue interpretable
models expressed symbolically as opposed to black-box neural networks,
the latter only being used during learning to efficiently parameterize
the large search space of possible models. In point (ii), learned
models are dynamically evaluated \emph{a posteriori} in the
computational solver instead of based on \emph{a priori} information
from preprocessed high-fidelity data, thereby accounting for the
specificity of the solver at hand such as its numerics. Finally in
point (iii), the exploration of new models is solely guided by
predefined integral quantities, e.g., averaged quantities of
engineering interest in Reynolds-averaged or large-eddy simulations
(LES). This also enables the assimilation of sparse data from
experimental measurements, which usually provide an averaged
large-scale description of the system rather than a detailed small-scale description. We use a coupled deep reinforcement learning
framework and computational solver to concurrently achieve these
objectives. The combination of reinforcement learning with objectives
(i), (ii) and (iii) differentiate our work from previous modeling
attempts based on machine learning.

The rest of this brief is organized as follows. In Section~2, we
provide a high-level description of the model discovery framework with
reinforcement learning. In Section~3, the method is detailed for the
application of discovering missing terms in differential equations. An
elementary instantiation of the method is described that discovers
missing terms in the Burgers' equation. Concluding remarks are offered
in Section~4.

\section{Model Discovery with Reinforcement Learning (MDRL)}
\label{sec:method}

\subsection{Offload human thinking by machine learning}

\begin{figure}
    \centering
    \vspace{5px}
    \includegraphics[clip,angle=-90,width=\textwidth]{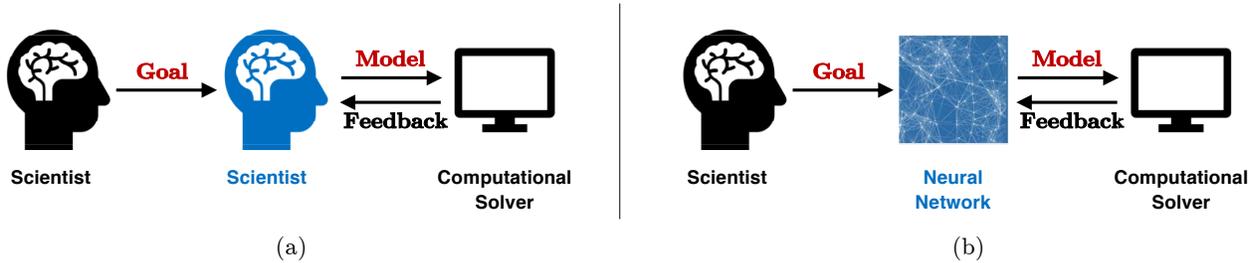}
    \caption{\label{fig:abstract} Iterative computational modeling process based on human intelligence (a) alone and (b) aided by artificial intelligence to offload human thinking during the iterative modeling process. In approach (b), neural network parameterizations allow to efficiently explore a large number of models with minimum bias. Blue colored elements highlight differences between both approaches. \label{fig:1}}
\end{figure}

The purpose of modeling is to devise a computational \emph{strategy}
to achieve a prescribed engineering or physical \emph{goal}, as
depicted in Figure~\ref{fig:1}. For example, we may search a
subgrid-scale (SGS) model for LES (strategy) that can accurately
predict the average mass flow in a pipe (goal). While it is natural
that determining the goal is a completely human oriented task, as we aim to solve a
problem of our own interest, there is no obvious reason why this
should be the case for the strategy.

Traditionally, as shown in Figure~\ref{fig:1}(a), finding a modeling
strategy heavily relies on human thinking, which encompasses
phenomenological understanding, physical insights, and mathematical
derivations, among others. Deriving a model is typically an iterative
process that involves testing and refining ideas
sequentially. Although strategies devised by human intelligence alone
have shown success in the past, many earlier models can be
substantially improved, and many remain to be found. Current modeling
strategies may be hindered by the limits of human cognition, for
example by a researcher's preconceived ideas and biases. In a sense,
we have a vast knowledge about traditional models but very limited
idea about what we really seek: breakthrough models. This often
constrains us to \emph{exploit} prior knowledge for convenience more
than we \emph{explore} innovative ideas. Following the example in the
previous paragraph, we can constrain the functional form of the SGS
model in LES to be an eddy viscosity model and focus on the eddy
viscosity parameter alone. This approach certainly facilitates the
strategy search as the space of all possible models is often too large
for researchers to exhaustively explore all of them. However, it is
perhaps at the expense of constraining the final strategy to a
suboptimal solution from the beginning of the modeling process.

Ideally, we would like to offload all human thinking of strategies to
artificial intelligence to efficiently explore the phase space of
models with minimum human bias, as sketched in Figure~1(b). Following
the previous example, our goal is to predict the average mass flow,
but whether we use LES or another computational approach, broadly defined, needs not be
decided by a human, but instead by an artificial intelligent agent. To
automate the full strategy process pertains to the bigger quest of
general artificial intelligence \citep{goertzel2007artificial}. The
aim of the present preliminary study is more modest. Yet, it is our
premise that current machine learning tools already enable to automate
a significant portion of the modeling strategy search. In the previous example
where we seek a model to estimate the pipe mass flow, we might constrain the strategy to employ a LES framework, then use the
artificial intelligent agent to specifically find the SGS model.

\subsection{Hybrid human/machine method based on reinforcement learning}

We pursue a method in which a reinforcement learning (RL) agent is
used to partially replace human thinking in devising strategies. RL
emulates how humans naturally learn and, similarly, how scientists
iterate during the modeling process. In particular, we draw
inspiration from the recent success of RL in achieving superhuman
performance across a number of tasks such as controlling robots
\citep{OpenAIDexterous} or playing complicated strategy games
\citep{silver2017mastering,silver2018general,alphastarblog}.

RL consists of training a software agent to take actions that maximize
a predefined notion of cumulative reward in a given environment
\citep{sutton2018reinforcement}: the RL \emph{agent} selects an
\emph{action} based on its current \emph{policy} and state, and sends
the action to the \emph{environment}. After the action is taken, the
RL agent receives a \emph{reward} from the environment depending on
the success in achieving a prescribed goal. The agent updates the
policy parameters based on the taken action and the reward
received. This process is continuously repeated during training,
allowing the agent to learn an optimal policy for the given
environment and reward signal.

\begin{figure}
    \centering
    \vspace{5px}
    \includegraphics[clip,angle=90,width=0.75\textwidth]{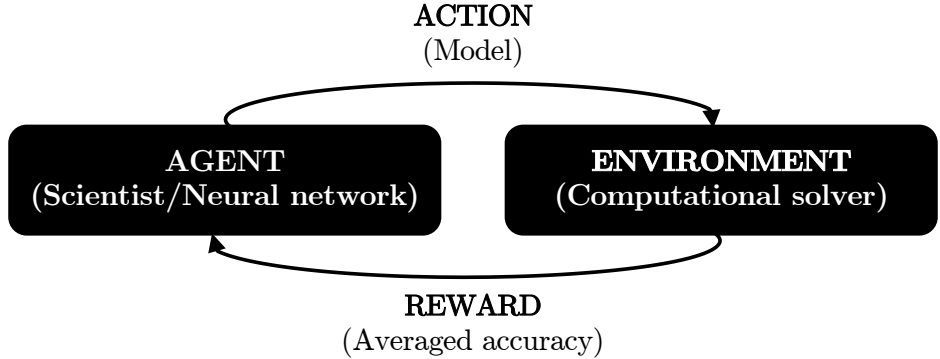}
    \caption{\label{fig:RL} Schematic representation of MDRL based
      on a reinforcement learning framework that mimics the scientific
      modeling process.}
\end{figure}

The MDRL methodology proposed here relies on a tailored RL framework
that mimics the scientific modeling process, as illustrated in
Figure~\ref{fig:RL}. MDRL consists of training a software scientist
(agent with a given policy) to iteratively search for the optimal
strategy or model (action) that maximizes the prescribed goals
(reward) obtained by using the model in a computational solver
(environment). As the state typically encountered in RL does not play
a role here, MDRL may be considered an example of the multi-armed
bandit problem \citep{sutton2018reinforcement}.

\section{Application of MDRL for analytical model discovery}
\label{sec:application}

Hereafter, we narrow the term model to refer to mathematical
expressions.  Instead of hand-designing new models from scratch, we
design a RL agent that searches for analytical formulas among the
space of known primitive functions.  In Section~3.1, we describe how
MDRL automates the process of discovering models in analytical
form. An elementary instantiation of the method is presented in
Section~3.2, where MDRL is utilized to efficiently discover missing
terms in the Burgers' equation.


\subsection{Description of the method}

\begin{figure}
    \centering
    \includegraphics[clip,angle=90,width=0.9\textwidth]{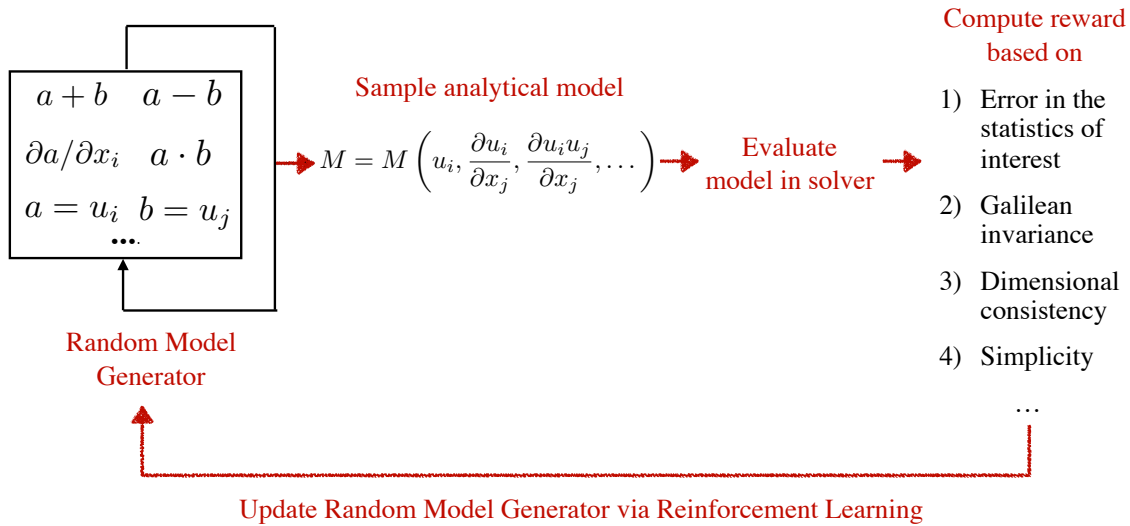}
    \caption{Schematic of the Model Discovery with Reinforcement
      Learning (MDRL) applied to the analytical discovery of models. A
      mathematical expression for the model is discovered following
      the requirements imposed as a ``reward''. For example, a typical
      reward would be a measure of global accuracy. Additional
      requirements can be enforced such as dimensional consistency,
      simplicity of the model, Galilean invariance,
      etc. \label{fig:AMD}}
\end{figure}

The workflow of MDRL for analytical model discovery is illustrated in
Figure~\ref{fig:AMD}.  The steps are summarized as follows:
\begin{enumerate}
    \item[(1)] A random model generator (RMG) is the agent that
      outputs mathematical expressions (actions) with a given
      probability distribution, similarly to a sequence of numbers
      generated by a random number generator. The RMG comprises a
      computational graph to encode mathematical formulas of arbitrary
      complexity in a particular domain-specific language (DSL).  The
      probabilities defining the RMG are parameterized as a neural
      network, which represents the policy of the RL agent.
      \item[(2)] A set of models is sampled from the RMG and decoded into
        their corresponding mathematical expressions ($M$ in
        Figure~\ref{fig:AMD}).
    \item[(3)] The sampled models are evaluated in the real
      environment using a computational solver.
    \item[(4)] The reward signal is calculated based on the prescribed goals
      and used to optimize the parameters (probabilities) of the RMG
      via a particular RL algorithm.
\end{enumerate}
The process described in Steps 1 to 4 is repeated until a convergence
criterion is satisfied.

The framework provided by MDRL has the unique feature of allowing the
researcher to focus on the intellectually relevant properties of the
model without the burden of specifying a predetermined analytical
form. The final outcome is an equation modeling the phenomena of
interest that can be inspected and interpreted, unlike other
widespread approaches based on neural networks. Our method generates a
mathematical equation, but could be similarly applied to generate
symbolic expressions for numerical discretizations.

We discuss next some of the components of MDRL for analytical model
discovery.

{\bf Domain-specific language}: A Domain-specific language (DSL) is
used to represent mathematical expressions in a form that is
compatible with the reinforcement learning task.  The purpose of the
DSL is to map models to actions in the RL framework.  Mathematical
expressions can be generally represented as computational graphs,
where the nodes correspond to operations or variables. Compatible
representations of these computational graphs can employ sequence-like
or tree-like structures \citep{bello2017neural, luo2018automatic,
  lample2019deep}, with varying degree of nesting depending on the
requested complexity of the sought-after mathematical expression.  It
is important to remark that the search space is not restricted to
elementary physical processes as the DSL allows for complex
combinations of elementary operations to build new ones, similarly as
to how with a few letters one can write a large number of words.  The
choice of the DSL plays an important role as it conditions the search
space and therefore the learning process.

{\bf Random model generator as a neural network}: The key insight that
enables the RL agent to efficiently learn a model in the vast search
space is to implicitly parameterize the latter using a neural
network. The RMG (agent) employs a stochastic policy network to decide
what models (actions) to sample. Actions are sampled according to a
multidimensional probability distribution $\pi_\theta$, where $\pi$ is
the neural network policy and $\theta$ its parameters. The objective
is to train the RMG to output models with increasing accuracy by
learning a probability distribution function $\pi_\theta$ that
maximizes the probability of sampling accurate models.
%
The RL training objective can be formulated as maximizing
\begin{equation}
J(\theta) = \mathbb{E}_{a\sim \pi_\theta (\cdot)}[R(a)],
\end{equation}
where $R(a)$ is the averaged accuracy (reward) of an action $a$ ($\equiv M$),
sampled according to a probability distribution $\pi_\theta$. The
objective is to maximize the expected performance $J(\theta)$ obtained
when sampling a model from the probability distribution outputted by
the policy. This is achieved by gradient ascending on the performance
objective $J$, using policy gradient algorithms for example.

A distributed training scheme can be employed to speed up the training
of the RMG. As the model evaluation is generally the most
time-consuming bottleneck, it is desirable to evaluate generated
models in parallel on distributed CPUs. In this manner, at each
iteration, the RMG samples a batch of models that are all run
simultaneously and later combined to update the probabilities of the
RMG.


\subsection{Example: discovering missing terms in the Burgers' equation}

\begin{figure}
    \centering
    \vspace{10px}
    \includegraphics[clip,angle=90,width=0.7\textwidth]{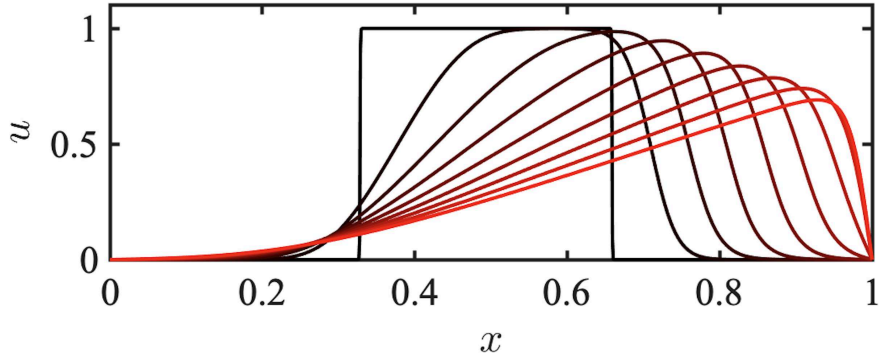}
    \caption{History of the solution to the Burgers' equation. \label{fig:burgers}}
\end{figure}

We apply the scheme proposed in Section~3.1 to discover the analytic form
of missing terms in a partial differential equation. We choose the
Burgers' equation as representative of the key features of a simple
fluid system,
\begin{equation}\label{eq:burgers}
\frac{\partial u}{\partial t} = -u\frac{\partial u}{\partial x} + \nu \frac{\partial^2 u}{ \partial x^2},
\end{equation}
where $u$ denotes velocity, $x$ and $t$ are the spatial and temporal
coordinates, respectively, and $\nu=0.01$ is the viscosity. We
simulate Eq.~(\ref{eq:burgers}) using as a initial condition a
rectangular signal. The equation is integrated from $t=0$ to $t=0.8$
using a fourth-order Runge-Kutta scheme for time-stepping and a
fourth-order central finite difference scheme for approximating the
spatial derivatives with 1000 points uniformly distributed in $x$. The
velocity profile is plotted at various time instants in
Figure~\ref{fig:burgers}.

The problem is formulated by re-arranging Eq.~(\ref{eq:burgers}) as
\begin{equation}\label{eq:burgersM}
  \frac{\partial u}{\partial t} =
  -\frac{1}{2}u\frac{\partial u}{\partial x} + \nu \frac{\partial^2 u}{ \partial x^2} +M, 
\end{equation}
where $M$ is an unknown mathematical expression that we aim to
discover using the MDRL framework proposed in
Figure~\ref{fig:AMD}. For each model $M=M(u,x,t,\partial u/\partial
x,...)$ sampled by the RMG, we compute the associated reward as
\begin{equation}
    R_i = \frac{1}{||u_{\mathrm{exact}}(x,T)-u_{\mathrm{model},i}(x,T)||+\epsilon} + \frac{\epsilon^{-1}}{n},
\end{equation}
where $||\cdot||$ is the L$_2$-norm, $u_{\mathrm{exact}}$ is the exact
solution at $t=T=0.8$, $u_{\mathrm{model},i}$ is the solution for the
$i$-th model at $t=T=0.8$, $n$ is the number of terms in $M$ (for
example $n=2$ for $M= u^2+\partial u/\partial x$), and
$\epsilon=0.1$. The first term in $R$ evaluates the accuracy of the
model, whereas the second term penalizes the model based on its
complexity (number of terms).  In each iteration of the RL process,
the RMG samples $m=100$ models, which results in the total
reward $\sum_{i=1}^m R_i$. The reward is then used to improve the RMG.

We discuss next the implementation of the RMG and the neural network
architecture for the RL agent used in this particular example. Both
the proper design of the RMG agent along with the choice of the
optimization method plays a major role in the success of the proposed
methodology. The specific choices made here have shown acceptable
performance, but we do not imply that these are optimal or extensible
to other cases.

For the computational graph representation of $M$, we follow a DSL
methodology similar to that used by \citet{bello2017neural}. Models
$M$ are formed by combining \emph{operands}, \emph{unary functions},
and \emph{binary functions}. Each group is composed of a few
elementary elements:
\begin{itemize}
\item \emph{Operands}: $u$, $x$, $t$, and integers $c$ from 1 to 100
  and the reciprocals $1/c$.
  \item{\emph{Unary functions}: $(\cdot)$ (identity), $-(\cdot)$ (sign
    flip), $\exp(\cdot)$, $\log(|\cdot|)$, $\sin(\cdot)$,
    $\cos(\cdot)$, and $\partial(\cdot)/\partial x$
    (differentiation).}
  \item \emph{Binary functions}: $+$ (addition), $-$ (subtraction),
    $\times$ (multiplication), and $/$ (division).
\end{itemize}
Formulas with an arbitrary number of terms are generated following a
recursive scheme similar to the one depicted in
Figure~\ref{fig:AMD}. The details of the specific DSL strategy adopted
here can be cumbersome. They are not emphasized as they are merely
designed for the particular showcase discussed in this
example. Further details regarding the unique and efficient
representation of analytical formulas using computational graphs can
be found in \citet{bello2017neural}, \citet{luo2018automatic}, and
\citet{lample2019deep}.

We use Deep Deterministic Policy Gradients (DDPG)
\citep{Lillicrap2015} as the network architecture for the RL
agent. The DDPG actor-critic algorithm is well suited for problems
with continuous action spaces, which we assimilate to the
probabilities required for the RMG in the current setting. The DDPG is
implemented using MATLAB (R2019a, The MathWorks Inc.) with default
parameters. The actor (RMG agent) is a multilayer perceptron with two
blocks, each with 5 fully connected hidden layers. Rectified linear
units and sigmoid activations are used in the first and second blocks,
respectively. The critic neural network is a multilayer perceptron
with 5 fully connected hidden layers with rectified linear units as
activation functions. Each layer contains roughly one hundred neurons.

\begin{figure}
    \centering
    \includegraphics[clip,width=0.8\textwidth]{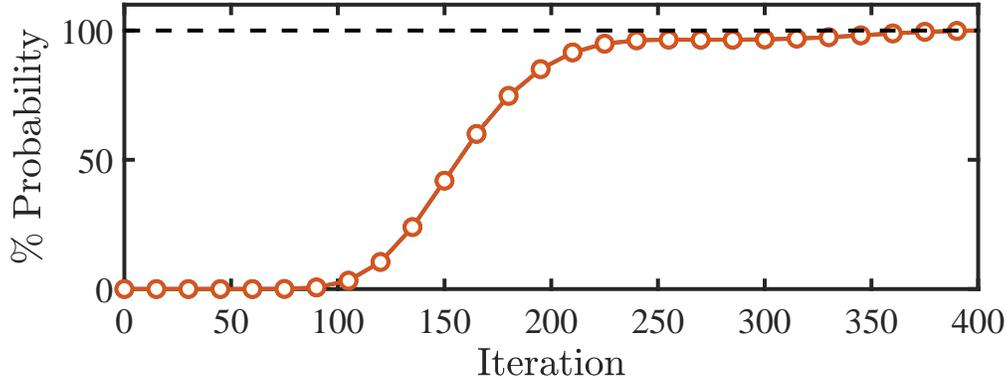}
    \caption{The probability of finding the exact solution for the
      missing term, $M=-1/2 \ u \ \partial u/\partial x$, as a
      function of the iteration number. The dashed line represents
      100\% probability of finding the exact
      solution. \label{fig:training}}
\end{figure}

The probability of finding the exact solution, $M=-1/2 \ u \ \partial
u/\partial x$, during the learning process is shown in
Figure~\ref{fig:training}. The result is obtained by performing 100
independent learning processes starting from scratch (RMG with uniform
probability distribution). After approximately 220 iterations, the
probability of discovering the exact solution is 99\%.
%
Our results can be compared with a random search approach, i.e. attempting to discover the model $M$ according to a fixed RMG with uniform probability distribution. The
latter typically requires more than $\mathcal{O}(10^9)$ iterations to find the
exact solution. Note that this random search approach was possible in the present example, but for cases with a vastly larger phase space, as in real-world applications, the random search becomes intractable and is likely to fail in finding an accurate model.  Hence, the results suggest that
MDRL is successful in detecting useful patterns in mathematical expressions and thereby speeds up the search process.

\section{Conclusions}
\label{sec:conclusions}

In this preliminary work, we discuss a paradigm for discovering reduced-order computational models assisted by
deep reinforcement learning. Our main premise is that state-of-the-art artificial intelligence algorithms enable to offload a significant portion of human thinking during the modeling process, in a manner that differs from the conventional approaches followed up to date.

We have first divided the modeling process into two broad tasks: defining goals and searching for strategies. Goals essentially summarize the problem we pursue to solve, whereas strategies denote practical solutions employed to achieve these goals.
Within the framework discussed here, we have argued that defining goals ought to remain in the realm of human thinking, as we aim to solve a problem of our interest. In contrast, strategies are mere practical means to achieve these goals. Thus we have advocated for a flexible methodology that intelligently search for optimal strategies while reducing scientists' biases during the search process. We deemed the latter necessary to prevent misleading preconceptions from hindering our potential to discover ground-breaking models.

In this brief, we presented the main ideas behind a hybrid human/AI method for model discovery by reinforcement learning (MDRL). The approach consists of learning models that are evaluated \emph{a posteriori} using exclusively integral quantities from the reference solution as prior knowledge. This allows the use of a wide source of averaged/large-scale computational and experimental data. The workflow is as follows. The scientist sets the goals, which defines the reward for the reinforcement learning agent. An intelligent agent devises a strategy that ultimately results in a model candidate. The model is evaluated in the real environment. The resulting performance, as measured by the reward, is fed back into the reinforcement learning agent that learns from that experience. The process is repeated until the model proposed by the agent meets the prescribed accuracy requirements. In the long term, it would be desirable to offload all human thinking of strategies to artificial intelligence. In this preliminary work, only a portion of the strategy is discovered by
machine learning.

We have detailed the scheme above by combining MDRL with computational graphs to learn analytical mathematical expressions. The approach comprises two components: a random model generator (agent) that generates symbolic expressions (models) according to a parameterized probability distribution, and a reinforcement learning algorithm that updates the parameters of the model generator. At each iteration, the agent learns to output better models by updating its parameters based on the error incurred by the tested models. As an example, we have applied MDRL to discovering missing terms in the Burgers' equation. This simple yet meaningful example shows how our approach retrieves the exact analytical missing term in the equation and outperforms a blind random search by several orders of magnitude in terms of computational cost.

Although the main motivating examples in this brief pertain to the field of fluid mechanics, the MDRL method can be applied to devising computational models or discovering theories in other engineering and scientific disciplines. Finally, the present brief should be understood as a statement of concepts and ideas, rather than a collection of best practices regarding the particular implementations (computational graphs, neural network architectures, etc.) of the MDRL for general problems. Refinement and further assessment of the method are currently investigated and will be discussed in future work.

\section*{Acknowledgments}

A.L.-D. acknowledges the support of NASA under grant No. NNX15AU93A
and of ONR under grant No. N00014-16-S-BA10. We thank Irwan Bello for
useful discussions. We also thank Jane Bae for her comments on the
brief.

\bibliographystyle{unsrt}

\begin{thebibliography}{30}
\expandafter\ifx\csname natexlab\endcsname\relax\def\natexlab#1{#1}\fi

\bibitem[Atkinson {\em et~al.\/}(2019)Atkinson, Subber, Wang, Khan, Hawi \&
  Ghanem]{atkinson2019data}
{\sc Atkinson, S., Subber, W., Wang, L., Khan, G., Hawi, P. \& Ghanem, R.} 2019
  Data-driven discovery of free-form governing differential equations. In {\em
  Advances in neural information processing systems (Second Workshop on Machine
  Learning and the Physical Sciences)\/}.

\bibitem[Bar-Sinai {\em et~al.\/}(2019)Bar-Sinai, Hoyer, Hickey \&
  Brenner]{bar2019learning}
{\sc Bar-Sinai, Y., Hoyer, S., Hickey, J. \& Brenner, M.~P.} 2019 Learning
  data-driven discretizations for partial differential equations. {\em
  Proc. Natl. Acad. Sci.\/} {\bf 116},
  15344--15349.

\bibitem[Bello {\em et~al.\/}(2017)Bello, Zoph, Vasudevan \&
  Le]{bello2017neural}
{\sc Bello, I., Zoph, B., Vasudevan, V. \& Le, Q.~V.} 2017 Neural optimizer
  search with reinforcement learning. In {\em Proceedings of the 34th
  International Conference on Machine Learning-Volume 70\/}, pp. 459--468.
  JMLR.org.

\bibitem[Bongard \& Lipson(2007)]{bongard2007automated}
{\sc Bongard, J. \& Lipson, H.} 2007 Automated reverse engineering of nonlinear
  dynamical systems. {\em Proc. Natl. Acad. Sci.\/}
  {\bf 104}, 9943--9948.

\bibitem[Brunton {\em et~al.\/}(2019)Brunton, Noack \&
  Koumoutsakos]{brunton2019machine}
{\sc Brunton, S.~L., Noack, B.~R. \& Koumoutsakos, P.} 2019 Machine learning
  for fluid mechanics. {\em Annu. Rev. Fluid Mech.\/} {\bf 52}.

\bibitem[Brunton {\em et~al.\/}(2016)Brunton, Proctor \&
  Kutz]{brunton2016discovering}
{\sc Brunton, S.~L., Proctor, J.~L. \& Kutz, J.~N.} 2016 Discovering governing
  equations from data by sparse identification of nonlinear dynamical systems.
  {\em Proc. Natl. Acad. Sci.\/} {\bf 113},
  3932--3937.

\bibitem[Chen {\em et~al.\/}(2018)Chen, Rubanova, Bettencourt \&
  Duvenaud]{chen2018neural}
{\sc Chen, T.~Q., Rubanova, Y., Bettencourt, J. \& Duvenaud, D.~K.} 2018 Neural
  ordinary differential equations. In {\em Adv. Neural Inf. Process Syst.\/}, pp. 6571--6583.

\bibitem[Esteva {\em et~al.\/}(2017)Esteva, Kuprel, Novoa, Ko, Swetter, Blau \&
  Thrun]{esteva2017dermatologist}
{\sc Esteva, A., Kuprel, B., Novoa, R.~A., Ko, J., Swetter, S.~M., Blau, H.~M.
  \& Thrun, S.} 2017 Dermatologist-level classification of skin cancer with
  deep neural networks. {\em Nature\/} {\bf 542}, 115.

\bibitem[Goertzel \& Pennachin(2007)]{goertzel2007artificial}
{\sc Goertzel, B. \& Pennachin, C.} 2007 {\em Artificial general
  intelligence\/}, vol.~2. Springer.

\bibitem[Jimenez(2020)]{Jimenez2020}
{\sc Jim\'enez, J.} 2020 Computers and turbulence. {\em Eur. J. Mech. B-Fluid\/} {\bf 79}, 1--11.

\bibitem[Lample \& Charton(2019)]{lample2019deep}
{\sc Lample, G. \& Charton, F.} 2019 Deep learning for symbolic mathematics.

\bibitem[Lee {\em et~al.\/}(2019)Lee, Kooshkbaghi, Spiliotis, Siettos \&
  Kevrekidis]{lee2019coarse}
{\sc Lee, S., Kooshkbaghi, M., Spiliotis, K., Siettos, C.~I. \& Kevrekidis,
  I.~G.} 2019 {Coarse-scale PDEs from fine-scale observations via machine
  learning}. {\em arXiv preprint arXiv:1909.05707\/}.

\bibitem[Lillicrap {\em et~al.\/}(2016)Lillicrap, Hunt, Pritzel, Heess, Erez,
  Tassa, Silver \& Wierstra]{Lillicrap2015}
{\sc Lillicrap, T.~P., Hunt, J.~J., Pritzel, A., Heess, N., Erez, T., Tassa,
  Y., Silver, D. \& Wierstra, D.} 2016 Continuous control with deep
  reinforcement learning. In {\em ICLR\/} (ed. Y.~Bengio \& Y.~LeCun).

\bibitem[Long {\em et~al.\/}(2019)Long, Lu \& Dong]{long2019pde}
{\sc Long, Z., Lu, Y. \& Dong, B.} 2019 {PDE-Net 2.0: Learning PDEs from data
  with a numeric-symbolic hybrid deep network}. {\em J. Comput. Phys.\/} {\bf 399}, 108925.

\bibitem[Long {\em et~al.\/}(2018)Long, Lu, Ma \& Dong]{long2018pde}
{\sc Long, Z., Lu, Y., Ma, X. \& Dong, B.} 2018 {PDE-Net: Learning PDEs from
  Data}. In {\em International Conference on Machine Learning\/}, pp.
  3214--3222.

\bibitem[Luo \& Liu(2018)]{luo2018automatic}
{\sc Luo, M. \& Liu, L.} 2018 Automatic derivation of formulas using
  reforcement learning. {\em arXiv preprint arXiv:1808.04946\/}.

\bibitem[Nagpal {\em et~al.\/}(2019)Nagpal, Foote, Liu, Chen, Wulczyn, Tan,
  Olson, Smith, Mohtashamian, Wren {\em et~al.\/}]{nagpal2019development}
{\sc Nagpal, K., Foote, D., Liu, Y., Chen, P.-H.~C., Wulczyn, E., Tan, F.,
  Olson, N., Smith, J.~L., Mohtashamian, A., Wren, J.~H. {\em et~al.\/}} 2019
  Development and validation of a deep learning algorithm for improving gleason
  scoring of prostate cancer. {\em NPJ Digit. Med.\/} {\bf 2}, 48.

\bibitem[OpenAI {\em et~al.\/}(2018)OpenAI, Andrychowicz, Baker, Chociej,
  J{\'{o}}zefowicz, McGrew, Pachocki, Pachocki, Petron, Plappert, Powell, Ray,
  Schneider, Sidor, Tobin, Welinder, Weng \& Zaremba]{OpenAIDexterous}
{\sc OpenAI, Andrychowicz, M., Baker, B., Chociej, M., J{\'{o}}zefowicz, R.,
  McGrew, B., Pachocki, J.~W., Pachocki, J., Petron, A., Plappert, M., Powell,
  G., Ray, A., Schneider, J., Sidor, S., Tobin, J., Welinder, P., Weng, L. \&
  Zaremba, W.} 2018 Learning dexterous in-hand manipulation. {\em CoRR\/} {\bf
  abs/1808.00177}.

\bibitem[Perrault {\em et~al.\/}(2019)Perrault, Shoham, Brynjolfsson, Clark,
  Etchemendy, Grosz, Lyons, Manyika \& Niebles]{AIreport2019}
{\sc Perrault, R., Shoham, Y., Brynjolfsson, E., Clark, J., Etchemendy, J.,
  Grosz, B., Lyons, T., Manyika, J. \& Niebles, S. M. J.~C.} 2019 {The AI Index
  2019 Annual Report}. AI Index Steering Committee, Human-Centered AI
  Institute, Stanford University, Stanford, CA.

\bibitem[Raissi \& Karniadakis(2018)]{raissi2018hidden}
{\sc Raissi, M. \& Karniadakis, G.~E.} 2018 Hidden physics models: Machine
  learning of nonlinear partial differential equations. {\em J. Comput. Phys\/} {\bf 357}, 125--141.

\bibitem[Raissi {\em et~al.\/}(2017)Raissi, Perdikaris \&
  Karniadakis]{raissi2017physics}
{\sc Raissi, M., Perdikaris, P. \& Karniadakis, G.~E.} 2017 Physics informed
  deep learning (part ii): Data-driven discovery of nonlinear partial
  differential equations. {\em arXiv preprint arXiv:1711.10566\/}.

\bibitem[Raumviboonsuk {\em et~al.\/}(2019)Raumviboonsuk, Krause,
  Chotcomwongse, Sayres, Raman, Widner, Campana, Phene, Hemarat, Tadarati {\em
  et~al.\/}]{raumviboonsuk2019deep}
{\sc Raumviboonsuk, P., Krause, J., Chotcomwongse, P., Sayres, R., Raman, R.,
  Widner, K., Campana, B.~J., Phene, S., Hemarat, K., Tadarati, M. {\em
  et~al.\/}} 2019 Deep learning versus human graders for classifying diabetic
  retinopathy severity in a nationwide screening program. {\em NPJ Digit. Med.\/} {\bf 2}, 25.

\bibitem[Rudy {\em et~al.\/}(2017)Rudy, Brunton, Proctor \& Kutz]{rudy2017data}
{\sc Rudy, S.~H., Brunton, S.~L., Proctor, J.~L. \& Kutz, J.~N.} 2017
  Data-driven discovery of partial differential equations. {\em Sci. Adv.\/} {\bf 3}, e1602614.

\bibitem[Schaeffer(2017)]{schaeffer2017learning}
{\sc Schaeffer, H.} 2017 Learning partial differential equations via data
  discovery and sparse optimization. {\em Proc. R. Soc. Lond. A\/} {\bf 473},
  20160446.

\bibitem[Schmidt \& Lipson(2009)]{schmidt2009distilling}
{\sc Schmidt, M. \& Lipson, H.} 2009 Distilling free-form natural laws from
  experimental data. {\em Science\/} {\bf 324}, 81--85.

\bibitem[Silver {\em et~al.\/}(2018)Silver, Hubert, Schrittwieser, Antonoglou,
  Lai, Guez, Lanctot, Sifre, Kumaran, Graepel {\em
  et~al.\/}]{silver2018general}
{\sc Silver, D., Hubert, T., Schrittwieser, J., Antonoglou, I., Lai, M., Guez,
  A., Lanctot, M., Sifre, L., Kumaran, D., Graepel, T. {\em et~al.\/}} 2018 A
  general reinforcement learning algorithm that masters chess, shogi, and go
  through self-play. {\em Science\/} {\bf 362}, 1140--1144.

\bibitem[Silver {\em et~al.\/}(2017)Silver, Schrittwieser, Simonyan,
  Antonoglou, Huang, Guez, Hubert, Baker, Lai, Bolton {\em
  et~al.\/}]{silver2017mastering}
{\sc Silver, D., Schrittwieser, J., Simonyan, K., Antonoglou, I., Huang, A.,
  Guez, A., Hubert, T., Baker, L., Lai, M., Bolton, A. {\em et~al.\/}} 2017
  Mastering the game of go without human knowledge. {\em Nature\/} {\bf
  550}, 354.

\bibitem[Sutton \& Barto(2018)]{sutton2018reinforcement}
{\sc Sutton, R.~S. \& Barto, A.~G.} 2018 {\em Reinforcement learning: An
  introduction\/}. MIT press.

\bibitem[Vinyals {\em et~al.\/}(2019)Vinyals, Babuschkin, Chung, Mathieu,
  Jaderberg, Czarnecki, Dudzik, Huang, Georgiev, Powell, Ewalds, Horgan,
  Kroiss, Danihelka, Agapiou, Oh, Dalibard, Choi, Sifre, Sulsky, Vezhnevets,
  Molloy, Cai, Budden, Paine, Gulcehre, Wang, Pfaff, Pohlen, Yogatama, Cohen,
  McKinney, Smith, Schaul, Lillicrap, Apps, Kavukcuoglu, Hassabis \&
  Silver]{alphastarblog}
{\sc Vinyals, O., Babuschkin, I., Chung, J., Mathieu, M., Jaderberg, M.,
  Czarnecki, W., Dudzik, A., Huang, A., Georgiev, P., Powell, R., Ewalds, T.,
  Horgan, D., Kroiss, M., Danihelka, I., Agapiou, J., Oh, J., Dalibard, V.,
  Choi, D., Sifre, L., Sulsky, Y., Vezhnevets, S., Molloy, J., Cai, T., Budden,
  D., Paine, T., Gulcehre, C., Wang, Z., Pfaff, T., Pohlen, T., Yogatama, D.,
  Cohen, J., McKinney, K., Smith, O., Schaul, T., Lillicrap, T., Apps, C.,
  Kavukcuoglu, K., Hassabis, D. \& Silver, D.} 2019 {AlphaStar: Mastering the
  Real-Time Strategy Game StarCraft II}.
  \url{https://deepmind.com/blog/alphastar-mastering-real-time-strategy-game-starcraft-ii/}.

\bibitem[Wu \& Zhang(2019)]{wu2019learning}
{\sc Wu, Z. \& Zhang, R.} 2019 Learning physics by data for the motion of a
  sphere falling in a non-newtonian fluid. {\em Commun. Nonlinear Sci. Numer. Simul.\/} {\bf 67}, 577--593.

\end{thebibliography}

\end{document}